\documentclass{article}

     \usepackage[final]{neurips_2019}

\usepackage[utf8]{inputenc} 
\usepackage[T1]{fontenc}    
\usepackage{hyperref}       
\usepackage{url}            
\usepackage{booktabs}       
\usepackage{amsfonts}       
\usepackage{nicefrac}       
\usepackage{microtype}      
\usepackage{listings}
\usepackage[pdftex]{graphicx}
\usepackage{subfig}
\usepackage{balance}
\usepackage{comment}
\usepackage{authblk}
\usepackage{xcolor}
\usepackage{xspace}
\usepackage{wrapfig}

\newcommand{\eg}{\textit{e.g.}}

\newcommand{\mvfstrl}{\textsc{mvfst-rl}\xspace}
\newcommand{\mvfst}{\textsc{mvfst}\xspace}
\newcommand{\cwnd}{\textit{cwnd}\xspace}

\setcitestyle{numbers,square,citesep={,}}

\newcounter{trCounter}
\newif\iftrvar
\trvartrue
\iftrvar
\newcommand{\tim}[1]{{\small \color{blue}
    \refstepcounter{trCounter}\textsf{[TR]$_{\arabic{trCounter}}$:{#1}}}}
\newcommand{\seb}[1]{{\small \color{magenta}
    \refstepcounter{trCounter}\textsf{[SR]$_{\arabic{trCounter}}$:{#1}}}}

\newcommand{\alex}[1]{{\small \color{orange}
    \refstepcounter{trCounter}\textsf{[AHM]$_{\arabic{trCounter}}$:{#1}}}}
\else
\newcommand{\tim}[1]{}

\newcommand{\alex}[1]{}
\newcommand{\seb}[1]{}
\fi

\title{\mvfstrl: An Asynchronous RL Framework for Congestion Control with Delayed Actions}

\author[1]{Viswanath Sivakumar}
\author[1]{Olivier Delalleau}
\author[1,2]{Tim Rocktäschel}
\author[1]{Alexander H. Miller}
\author[1]{Heinrich Küttler}
\author[1,3]{Nantas Nardelli}
\author[1]{Mike Rabbat}
\author[1,4]{Joelle Pineau}
\author[1,2]{Sebastian Riedel}
\affil[1]{Facebook AI Research}
\affil[2]{University College London}
\affil[3]{University of Oxford}
\affil[4]{McGill University \& MILA}

\begin{document}

\maketitle

\begin{abstract}
Effective network congestion control strategies are key to keeping the Internet (or any large computer network) operational. Network congestion control has been dominated by hand-crafted heuristics for decades. Recently, Reinforcement Learning~(RL) has emerged as an alternative to automatically optimize such control strategies. Research so far has primarily considered RL interfaces which \emph{block} the sender while an agent considers its next action. This is largely an artifact of building on top of frameworks designed for RL in games (\eg\ OpenAI Gym). However, this does not translate to real-world networking environments, where a network sender waiting on a policy without sending data leads to under-utilization of bandwidth. We instead propose to formulate congestion control with an asynchronous RL agent that handles delayed actions. We present \mvfstrl, a scalable framework for congestion control in the QUIC transport protocol that leverages state-of-the-art in asynchronous RL training with off-policy correction. We analyze modeling improvements to mitigate the deviation from Markovian dynamics, and evaluate our method on emulated networks from the Pantheon benchmark platform. The source code is publicly available at \href{https://github.com/facebookresearch/mvfst-rl}{https://github.com/facebookresearch/mvfst-rl}.
\end{abstract}

\section{Introduction}

Congestion control is one of the key components of any large computer network, most notably the Internet, and is important to enable operation at scale.\footnote{It is estimated that 150,000 PB of data were sent per month over the Internet in 2018 (\url{https://www.statista.com/statistics/499431/global-ip-data-traffic-forecast/}).}
The goal of a congestion control algorithm is to dynamically regulate the rate of data being sent at each node to maximize total throughput and minimize queuing delay and packet loss.
The vast majority of network strategies still rely on hand-crafted heuristics that are reactive rather than predictive. An early method, called Remy~\cite{remy}, demonstrated that offline-learned congestion control can be competitive with conventional methods. More recently, RL-based approaches have been proposed and show promise in simulated environments~\cite{pcc-rl, iroko}.

\begin{wrapfigure}{R}{0.4\textwidth}
  \vspace{-15pt}    
  \begin{center}
    \includegraphics[width=0.38\textwidth]{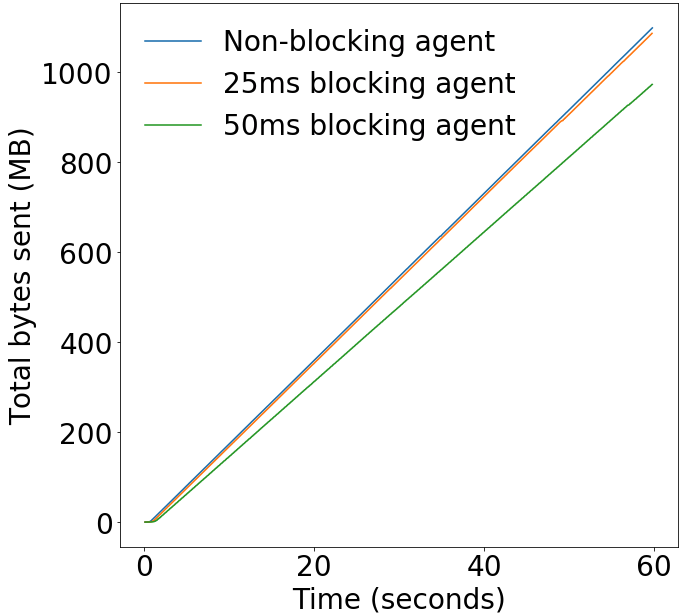}
  \end{center}
  \caption{Cumulative bytes sent during 60 seconds over AWS California to Mexico emulator. RL agents which block the sender for 25ms and 50ms waiting on the policy transmit 1.1\% and 11.4\% fewer bytes, respectively, compared to the non-blocking agent. All agents take actions at 100ms intervals.}
  \label{fig:blocking_agent}
  \vspace{-40pt}
\end{wrapfigure}

Despite the above advances, to our knowledge, no RL method has been transferred to real-world production systems. One of the biggest drawbacks of RL congestion control from a deployment perspective is that policy lookup takes orders of magnitude longer compared to hand-crafted methods. Moreover, networking servers often have limited resources to run a machine learning model, thus requiring inference to be offloaded to dedicated servers and further delaying action updates. Current RL congestion control environments follow the synchronous RL paradigm (e.g.\ using the OpenAI Gym~\cite{gym} interface), where model execution blocks the environment (network sender). This makes it infeasible for deployment where a sender waiting on a synchronous RL agent for congestion control, even for a few milliseconds, negatively impacts throughput (Figure~\ref{fig:blocking_agent}).

In this paper we introduce \mvfstrl, a training framework that addresses these issues with a non-blocking RL agent for congestion control. For training in the presence of asynchronous interaction between the environment and the learner (owing to the inability of the environment to wait for gradient updates), we leverage IMPALA~\cite{impala} which uses importance-sampling for off-policy correction. \mvfstrl is built on \mvfst\footnote{\href{https://github.com/facebookincubator/mvfst}{https://github.com/facebookincubator/mvfst}}, a C++ implementation of the QUIC transport protocol used in Facebook production networks, allowing seamless transfer to deployment. To emulate real-world network traffic in RL environments for training, we build upon Pantheon~\cite{pantheon} which obviates the need for hand-written traffic patterns and network topologies. We evaluate training with delayed actions and present our results on the Pantheon test-bed.

\section{Related Work}

A few different studies have applied RL to congestion control in varying ways. Iroko~\cite{iroko} takes the approach of a centralized policy to regulate sending rates of all the nodes in a network topology. While this is applicable for small to medium sized networks, it is intractable for large or Internet-scale networks. Iroko also requires manually specifying network topologies and traffic generators, making it difficult to emulate real-world network conditions. PCC-RL~\cite{pcc-rl} implements congestion control as an environment in OpenAI Gym~\cite{gym}, with a blocking RL agent. Park's~\cite{park} congestion control environment based on CCP~\cite{ccp} is closest to \mvfstrl in design with Remote Procedure Calls (RPC) for environment--agent communication, but it effectively takes a synchronous approach with its short step-time of 10ms, constraining to only very small models.

\section{Congestion Control as MDP with Delayed Actions}

Consider a Markov Decision Process (MDP)~\cite{mdp} formulated as \(\langle{}S, A, T, R\rangle{}\) where \(S\) is a set of states, \(A\) is a set of actions, \(R\) is the reward obtained, and \(T(s'|s,a)\) is the state-transition probability function. Unrolling over time, we have the trajectory \((s_t, a_t, r_t, s_{t+1}, a_{t+1}, r_{t+1}, ...)\). Casting congestion control as an MDP, the state space includes network statistics gathered from acknowledgement packets (ACKs) from the receiver such as round-trip time (RTT), queuing delay, packets sent, acknowledged and lost. The action is an update to the congestion window size, \cwnd, which is the number of unacknowledged packets that are allowed to be in flight. Large \cwnd leads to high throughput, but also increased delay due to queuing at intermediate buffers and network congestion. Our choice of action space \(A\) is a discrete set of updates to \cwnd: \(\{\cwnd, \cwnd/2, \cwnd-10, \cwnd+10, \cwnd\times 2\}\) (Appendix \ref{app:action_space}). The reward is generally a function of measured throughput and delay, aiming to strike a trade-off.

When an agent acts asynchronously on the environment at fixed time-intervals, the action \(a_t\) is applied after a delay of \(\delta\) such that \(t < t+\delta < t+1\), where \(\delta\) is the policy lookup time. The environment meanwhile would have transmitted further data based on the old action \(a_{t-1}\) during the interval \([t, t+\delta]\). The next state \(s_{t+1}\) therefore depends not just on \(s_t\) and \(a_t\), but also on \(a_{t-1}\). To incorporate this into the MDP, the state space can be augmented with the last action taken. In addition to this, given that our action space is relative to the previous actions taken, we find that a longer history of actions in the environment state helps learning. Therefore, we augment our state space with an action history of generic length \(h\), which we treat as a hyperparameter, and obtain our augmented state space \(\hat S = S \times A^h\). At time \(t\), the augmented state \(\hat s_t\) is \((s_t, a_{t-1},..., a_{t-h})\). With this formalism, we define our environment state and reward function below.

\textbf{Environment State:} Our choice of state vector contains a summary of network statistics gathered over a 100ms window along with a history of actions taken. From each ACK within the window, we gather 20 features such as the latest and minimum RTT, queuing delay, bytes sent, bytes acknowledged, and bytes lost and re-transmitted. The complete set of features are listed in Appendix \ref{app:state_space}. To normalize the features, the time fields are measured in milliseconds and scaled by $10^{-3}$ and byte fields are in KB\@. We follow Copa's~\cite{copa} strategy of computing \textit{RTTStanding} to reduce noise in delay measurements. For each feature, we calculate sum, mean, standard deviation, min and max within a window, and concatenate them with \(h\) one-hot vectors of past actions along with the \cwnd on applying those actions. This results in a state vector of size \(100 + h \times (|A| + 1)\), where \(A\) is the action space defined earlier.

\textbf{Reward:} Our reward is computed as $\log(t + \varepsilon) - \beta \log(d + \varepsilon)$ where \(t\) and \(d\) are the average throughput in MB/sec and the maximum delay in milliseconds, respectively, during the 100ms window, and the parameter $\beta$ trades-off between the two. $\varepsilon$ is a small value to ensure numerical stability.
We also experimented with a linear trade-off $t - \beta d$, but we found it difficult to find a value of $\beta$ that works well when trained jointly over different network scenarios where the values of $t$ and $d$ may vary wildly.
In such a setup (where $t$ and $d$ can be very different across the multiple scenarios the model is being trained on), even our log-based reward is affected by another issue: since its operational range may be quite different between the scenarios, the model may not treat all scenarios equally, favoring a few of them to maximize the average reward.
We address this issue with a reward normalization scheme by maintaining an online estimate of the mean and standard deviation of the reward in each scenario, which is used to normalize the reward during training.

In the following sections, we describe a framework for asynchronous RL training of congestion control with delayed actions and provide experimental analysis of the MDP with the augmented state space.

\section{Introducing \mvfstrl}

\begin{figure}
  \centering
  \subfloat[Asynchronous RL agent--sender interaction]{
    \includegraphics[width=6cm, keepaspectratio]{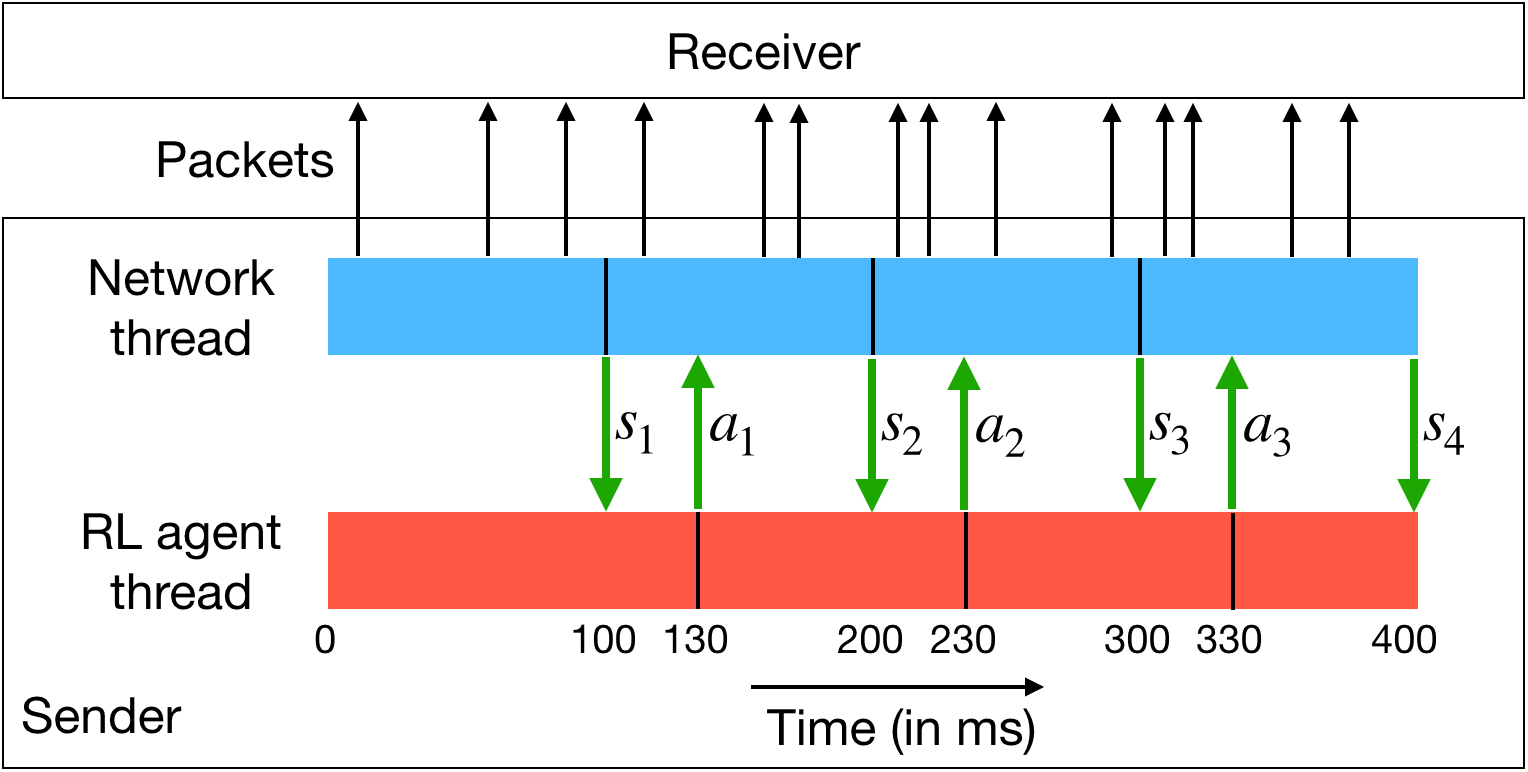}
    \label{fig:rl_agent}
    }%
  \qquad
  \subfloat[Training architecture]{
    \includegraphics[width=6.5cm, keepaspectratio]{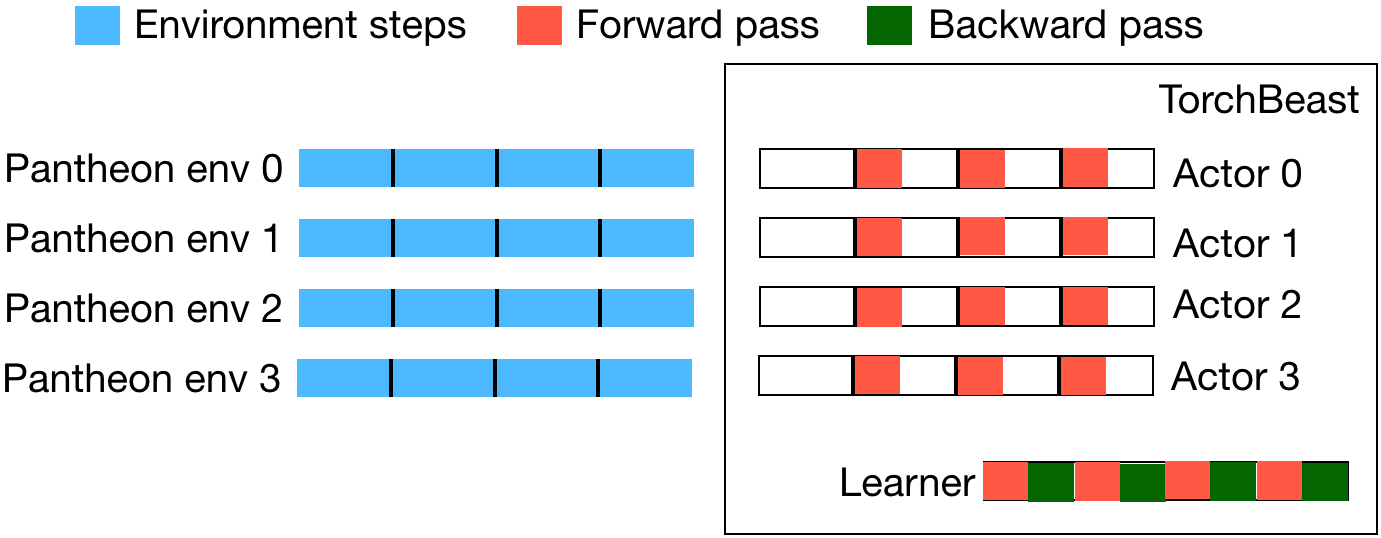}
    \label{fig:torchbeast_architecture}
    }%
  \caption{(a) States are sent to the agent every 100ms and in this illustration policy lookup takes 30ms (in practice in our experiments it is more around ~3ms). In the synchronous paradigm, packets could not be sent during the 30ms sections between \(s_t\) and \(a_t\). (b) TorchBeast maintains one actor per Pantheon environment, with RPC for state and action updates. The key difference from IMPALA is that, in addition to the actor--learner interaction, the environment--actor interaction is also asynchronous and the environment steps are not blocked by the forward-pass.}
  \label{fig:architecture}
  \vspace{-15pt}
\end{figure}

\textbf{Framework:} \mvfstrl follows an environment-driven design with an asynchronous RL agent. It integrates three components: (1) \mvfst, Facebook's implementation of the QUIC transport protocol used in production servers, (2) TorchBeast\footnote{\href{https://github.com/facebookresearch/torchbeast}{https://github.com/facebookresearch/torchbeast}}\cite{torchbeast}, a PyTorch~\cite{pytorch} implementation of IMPALA~\cite{impala} for fast, asynchronous, parallel RL training, and (3) Pantheon~\cite{pantheon} network emulators calibrated with Bayesian Optimization~\cite{bayesopt} to match real-world network conditions. The RL congestion controller accumulates network statistics from ACKs over a fixed time-window, say 100ms, and sends a state update asynchronously to an RL agent in a separate thread. Once the agent executes the policy, the action to update \cwnd is applied asynchronously to the transport layer (Figure~\ref{fig:rl_agent}).

IMPALA was originally designed to overcome scaling limitations during training of batched Actor-Critic methods, where parallel actors act for \(T\) steps each to gather trajectories (sequences of \(s_t, a_t, r_t, s_{t+1}\)), and then wait for a gradient update on the policy by the learner. IMPALA relaxes this restriction, and with an off-policy correction algorithm called V-trace, it allows actors to continue with trajectories while the learner asynchronously updates the policy. Although the original motivation was higher training throughput, the decoupled actor-learner algorithm and off-policy correction is well-suited for \mvfstrl, allowing the actors corresponding to network senders to not block on gradient updates, as doing so would change the dynamics of the underlying network.

Figure~\ref{fig:torchbeast_architecture} illustrates the training architecture. Each actor corresponds to a separate Pantheon instance in a sender--receiver setup with a randomly chosen emulated network scenario. The senders run the QUIC protocol with RL congestion controller and communicate state updates via RPC to the actors. The actors execute the model for the next action, and once a full trajectory is obtained, they communicate it to the learner for a gradient update. All communications are asynchronous and do not block the network sender. After training, the model can be exported with TorchScript and executed in C++ on the network host without the need for RPC.

Most RL optimization criteria (including V-trace used in IMPALA) rely on bootstrapping, i.e., in its simplest form, on the relation $V(s_t) = r_t + \gamma V(s_{t+1})$.
However, in an episodic setting, when $s_{t+1}$ is a terminal state
these criteria typically assume that the agent will get no additional reward,
and thus simply use $V(s_t) = r_t$.
This might be problematic given our training setup where the end of an episode occurs after an arbitrary amount of elapsed time~(30s),
as the critic could use its LSTM memory to anticipate the end of the episode to adjust the predicted state value accordingly.
Since the LSTM state is shared between the actor and the critic, there is thus a risk at deployment time for the LSTM state to evolve in unanticipated ways after the first 30s, which could affect the actor's behavior.
To address this concern, we follow the recommendation from~\cite{pardo2018time} to bootstrap the state value on episode termination, i.e., still use $V(s_t) = r_t + \gamma V(s_{t+1})$ even if  $s_{t+1}$ is a terminal state.
We implemented this logic in our custom version of V-trace.

\textbf{Model:} Our model trunk is a two-layer fully-connected neural network with 1024 units each and ReLU non-linearity. The extracted features and the (un-normalized) reward are fed into a single-layer LSTM~\cite{lstm}. For correctness during training, along with a trajectory of \(T\) steps, an actor also communicates the LSTM hidden state at the beginning of the trajectory to the learner. Finally, the LSTM output is fed into two heads -- a policy head with \(|A|\) outputs, and another head for the policy gradient baseline.

\section{Experiments}

\textbf{Training:} Our training setup consists of 40 parallel actors, each corresponding to a randomly chosen calibrated network emulator. We train episodically with each Pantheon sender--receiver instance running for 30 seconds, with the episodic trajectory used as a rollout for training. Episodic training provides the opportunity to learn dynamics similar to TCP Slow-Start, where the behavior of the algorithm during startup is sufficiently different from steady state. Following hyperparameter sweeps, we set the initial learning rate to $10^{-4}$ with RMSProp~\cite{rmsprop} and the trade-off parameter \(\beta\) in the reward to 0.75 ($\varepsilon$ is set to $10^{-5}$). All experiments are run for a total of 5M steps.

\begin{figure}
  \centering
  \subfloat[Nepal to AWS India (calibrated)]{
    \hspace*{10pt}%
    \includegraphics[width=8cm, keepaspectratio]{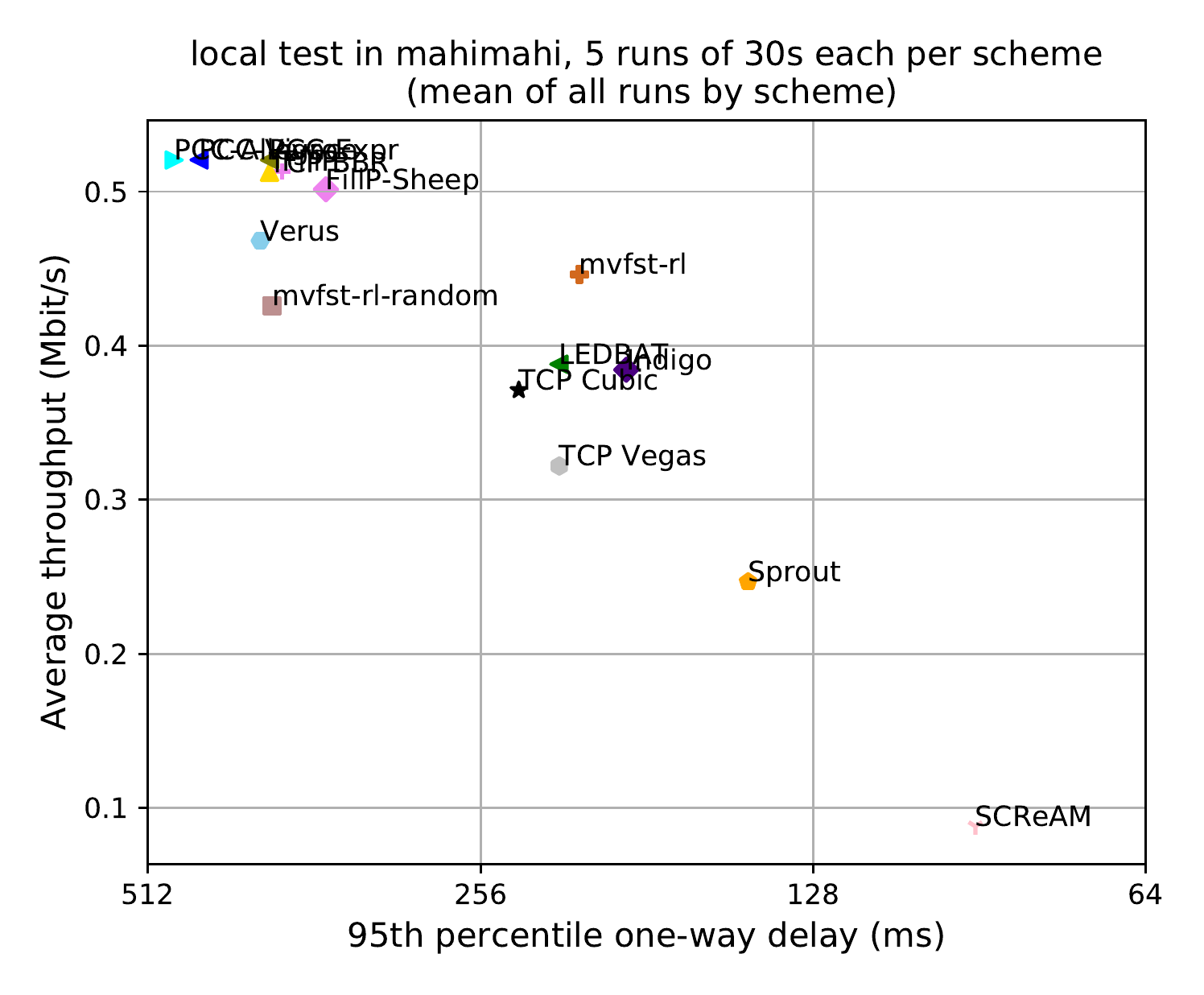}
    \label{fig:pantheon_job0}
    \hspace*{10pt}%
  }%
  \qquad
  \subfloat[Token-bucket policer (unseen environment)]{
    \hspace*{10pt}%
    \includegraphics[width=8cm, keepaspectratio]{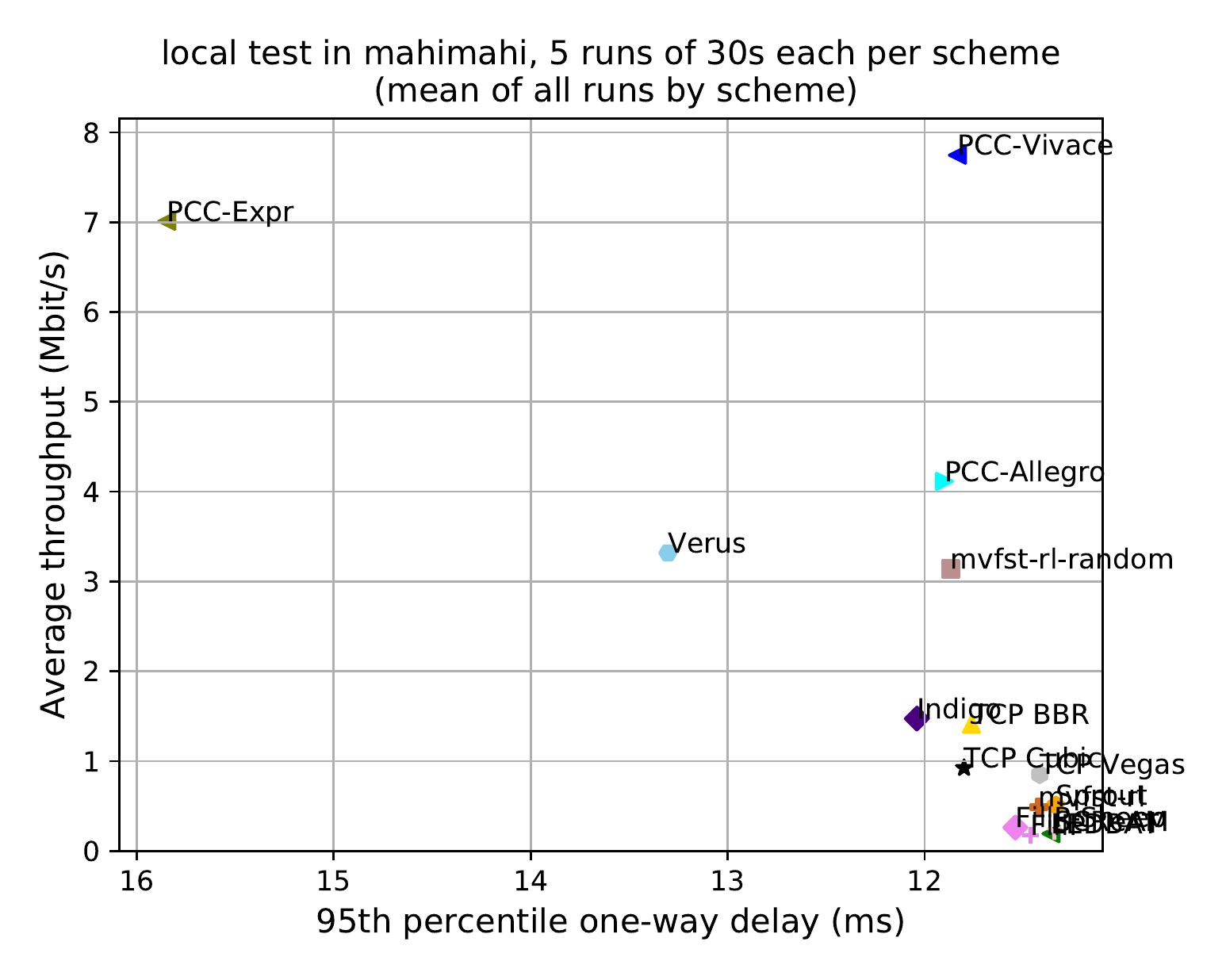}
    \label{fig:pantheon_job8}
    \hspace*{10pt}%
  }%
  \qquad

  \subfloat[With LSTM]{
    \includegraphics[width=4cm, keepaspectratio]{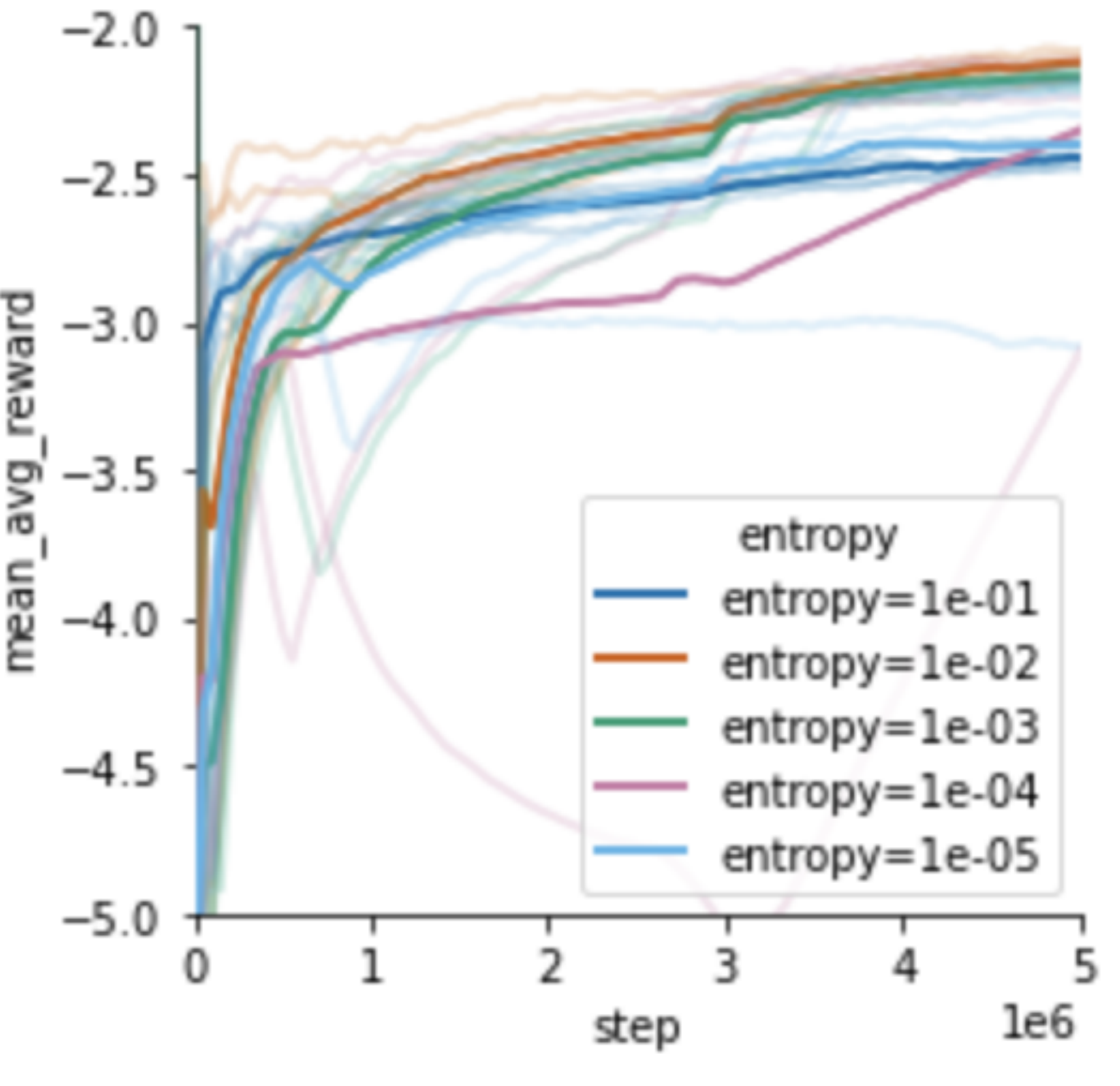}
    \label{fig:ablation_entropy_lstmTrue}
    }%
  \qquad
  \subfloat[Without LSTM]{
    \includegraphics[width=4cm, keepaspectratio]{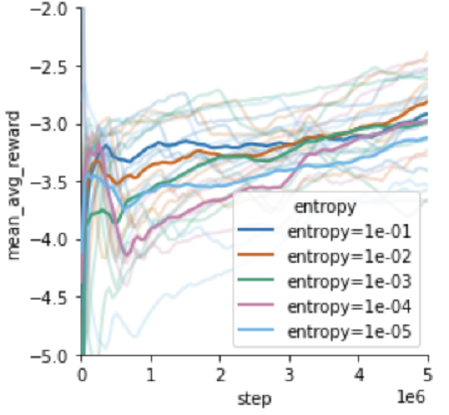}
    \label{fig:ablation_entropy_lstmFalse}
    }%
  \qquad
  \subfloat[Entropy penalty = 0.01]{
    \includegraphics[width=4cm, keepaspectratio]{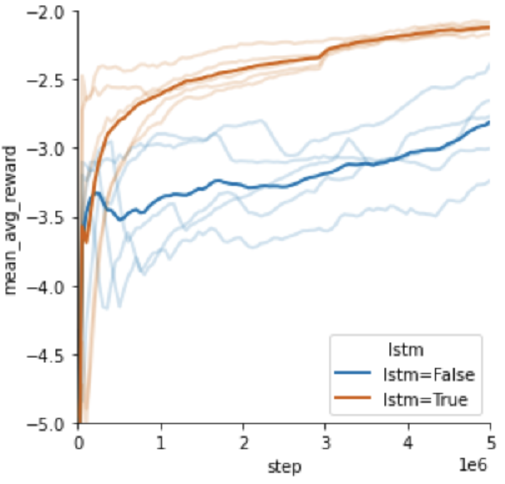}
    \label{fig:ablation_lstm}
    }%
  \caption{Experimental Results. (a) and (b):~Average throughput and 95th percentile delay when tested against calibrated Pantheon emulators. \texttt{rl-random} refers to a baseline agent that picks a random action over the same action space as \texttt{mvfst-rl} at each step. Mean of five 30-second runs. (c)~--~(e):~Training rewards with and without LSTM for varying values of entropy penalty.}
  \label{fig:experiments}
  \vspace{-10pt}
\end{figure}

\textbf{Results:} Figure~\ref{fig:pantheon_job0} shows the test performance on a Pantheon emulator calibrated against Nepal to AWS India. \mvfstrl (jointly trained over six network scenarios including this one) achieves a meaningful trade-off between throughput and delay, and appears competitive with other congestion control algorithms.
Compared to \texttt{rl-random}, a baseline agent that picks a random action over the same action space \(A\) at each step, our model achieves slightly higher throughput with significantly lower delay.
Overall we find that on the six training network scenarios, our trained model tends to achieve a lower throughput than the other congestion control algorithms with highest throughput, but also leads to a lower delay.
It should be possible in theory to tweak the reward (and in particular the trade-off coefficient $\beta$) to achieve higher throughput (at the cost of extra delay), but we have not yet investigated the full spectrum of solutions that may be obtained this way.

We also evaluate on a held-out network scenario, unseen during training, where our model does not perform as well (see Figure~\ref{fig:pantheon_job8}, where the $mvfst\_rl$ data point is among the cluster in the bottom right corner).
Note that this unseen network scenario is qualitatively different from the training ones, due to a combination of (i) lower RTT (20ms vs.~a minimum of 54ms in training scenarios) and (ii) using a token-bucket based policer which results in a much smaller buffer queue capacity (one single packet vs.~a minimum of fourteen packets in training scenarios).
Overall we find that the model does not generalize well to such new network scenarios unseen during training, which is not surprising given the limited amount of variability in training conditions, making it likely for the model to overfit to specific scenarios.
We leave to future work the problem of training more robust models, which could be achieved for instance by introducing domain randomization during training instead of using a fixed set of scenarios.

\textbf{Ablations:} We examine in Figure~\ref{fig:ablation_entropy_lstmTrue} the impact of IMPALA's entropy regularization coefficient on the total reward obtained per episode, over the course of training (results are average over five seeds).
We observe that low entropy regularization can make training unstable (the agent may converge prematurely to a deterministic behavior), while if it is too high then performance is negatively affected (due to the agent behaving too randomly).
We chose $0.01$ as the optimal value for this hyper-parameter.
We also run the same experiment without LSTM and find the performance to be significantly degraded (Figures \ref{fig:ablation_entropy_lstmFalse} and \ref{fig:ablation_lstm}). These results highlight the need to carefully tweak the entropy penalty, and the important role of recurrent connections (which we believe to be particularly helpful due to the partial observability of the environment). 

\section{Conclusion and Future Work}

RL is a promising direction for improving real-world systems. Yet, it is challenging to deploy RL in datacenters due to performance and resource constraints, especially when evaluated against hand-written heuristics. We take the first steps towards moving away from blocking RL agents for systems problems and introduce a framework for learning asynchronous RL policies in the face of delayed actions. When applied to congestion control, our initial results show promise in the MDP with augmented state space.
One challenge we face is learning a joint policy for a range of network scenarios when reward scales vary wildly due to differences inherent in the environment. We believe this is an interesting direction for further research on reward normalization strategies across environments: our current implementation based on per-environment statistics would not scale if we introduced more randomized scenarios (which will be required to improve the generalization ability of our approach).
We hope to evaluate our trained congestion control agents in production networks in the future by leveraging \mvfstrl's efficient implementation and tight integration with the QUIC transport protocol. \mvfstrl has been open-sourced and is available at \href{https://github.com/facebookresearch/mvfst-rl}{https://github.com/facebookresearch/mvfst-rl}.

\subsubsection*{Acknowledgments}
We would like to thank Jakob Foerster for fruitful discussions on applying RL for networks, Udip Pant, Ranjeeth Dasineni and Subodh Iyengar for their support related to \mvfst, Francis Cangialosi for insightful discussions around CCP, and Venkat Arun for sharing his congestion control wisdom.

\small

\medskip

\balance
\bibliographystyle{abbrv}
\bibliography{bibliography_doc}

\begin{thebibliography}{10}

\bibitem{copa}
V.~Arun and H.~Balakrishnan.
\newblock Copa: Practical delay-based congestion control for the internet.
\newblock In {\em 15th {USENIX} Symposium on Networked Systems Design and
  Implementation ({NSDI} 18)}, pages 329--342, Renton, WA, Apr. 2018. {USENIX}
  Association.

\bibitem{gym}
G.~Brockman, V.~Cheung, L.~Pettersson, J.~Schneider, J.~Schulman, J.~Tang, and
  W.~Zaremba.
\newblock Openai gym.
\newblock {\em CoRR}, abs/1606.01540, 2016.

\bibitem{impala}
L.~Espeholt, H.~Soyer, R.~Munos, K.~Simonyan, V.~Mnih, T.~Ward, Y.~Doron,
  V.~Firoiu, T.~Harley, I.~Dunning, S.~Legg, and K.~Kavukcuoglu.
\newblock {IMPALA}: Scalable distributed deep-{RL} with importance weighted
  actor-learner architectures.
\newblock In J.~Dy and A.~Krause, editors, {\em Proceedings of the 35th
  International Conference on Machine Learning}, volume~80 of {\em Proceedings
  of Machine Learning Research}, pages 1407--1416, Stockholmsmässan, Stockholm
  Sweden, 10--15 Jul 2018. PMLR.

\bibitem{rmsprop}
G.~Hinton, N.~Srivastava, and K.~Swersky.
\newblock Overview of mini-batch gradient descent, 2012.

\bibitem{lstm}
S.~Hochreiter and J.~Schmidhuber.
\newblock Long short-term memory.
\newblock {\em Neural Computation}, 9:1735--1780, 1997.

\bibitem{pcc-rl}
N.~Jay, N.~Rotman, B.~Godfrey, M.~Schapira, and A.~Tamar.
\newblock A deep reinforcement learning perspective on internet congestion
  control.
\newblock In K.~Chaudhuri and R.~Salakhutdinov, editors, {\em Proceedings of
  the 36th International Conference on Machine Learning}, volume~97 of {\em
  Proceedings of Machine Learning Research}, pages 3050--3059, Long Beach,
  California, USA, 09--15 Jun 2019. PMLR.

\bibitem{torchbeast}
H.~K\"{u}ttler, N.~Nardelli, T.~Lavril, M.~Selvatici, V.~Sivakumar,
  T.~Rockt\"{a}schel, and E.~Grefenstette.
\newblock {TorchBeast: A PyTorch Platform for Distributed RL}.
\newblock {\em arXiv preprint arXiv:1910.03552}, 2019.

\bibitem{park}
H.~Mao, P.~Negi, A.~Narayan, H.~Wang, J.~Yang, H.~Wang, R.~Marcus, R.~Addanki,
  M.~Khani, S.~He, V.~Nathan, F.~Cangialosi, S.~B. Venkatakrishnan, W.-H. Weng,
  S.~Han, T.~Kraska, and M.~Alizadeh.
\newblock Park: An open platform for learning augmented computer systems.
\newblock In {\em ICML Reinforcement Learning for Real Life Workshop}, 2019.

\bibitem{bayesopt}
J.~Mo{\v{c}}kus.
\newblock On bayesian methods for seeking the extremum.
\newblock In G.~I. Marchuk, editor, {\em Optimization Techniques IFIP Technical
  Conference Novosibirsk, July 1--7, 1974}, pages 400--404, Berlin, Heidelberg,
  1975. Springer Berlin Heidelberg.

\bibitem{ccp}
A.~Narayan, F.~Cangialosi, D.~Raghavan, P.~Goyal, S.~Narayana, R.~Mittal,
  M.~Alizadeh, and H.~Balakrishnan.
\newblock Restructuring endpoint congestion control.
\newblock In {\em Proceedings of the 2018 Conference of the ACM Special
  Interest Group on Data Communication}, SIGCOMM '18, pages 30--43, New York,
  NY, USA, 2018. ACM.

\bibitem{pardo2018time}
F.~Pardo, A.~Tavakoli, V.~Levdik, and P.~Kormushev.
\newblock Time limits in reinforcement learning, 2018.

\bibitem{pytorch}
A.~Paszke, S.~Gross, S.~Chintala, G.~Chanan, E.~Yang, Z.~DeVito, Z.~Lin,
  A.~Desmaison, L.~Antiga, and A.~Lerer.
\newblock Automatic differentiation in {PyTorch}.
\newblock In {\em NIPS Autodiff Workshop}, 2017.

\bibitem{mdp}
M.~L. Puterman.
\newblock {\em Markov Decision Processes: Discrete Stochastic Dynamic
  Programming}.
\newblock John Wiley \& Sons, Inc., New York, NY, USA, 1st edition, 1994.

\bibitem{iroko}
F.~Ruffy, M.~Przystupa, and I.~Beschastnikh.
\newblock Iroko: {A} framework to prototype reinforcement learning for data
  center traffic control.
\newblock {\em CoRR}, abs/1812.09975, 2018.

\bibitem{remy}
K.~Winstein and H.~Balakrishnan.
\newblock Tcp ex machina: Computer-generated congestion control.
\newblock In {\em Proceedings of the ACM SIGCOMM 2013 Conference on SIGCOMM},
  SIGCOMM '13, pages 123--134, New York, NY, USA, 2013. ACM.

\bibitem{pantheon}
F.~Y. Yan, J.~Ma, G.~D. Hill, D.~Raghavan, R.~S. Wahby, P.~Levis, and
  K.~Winstein.
\newblock Pantheon: the training ground for internet congestion-control
  research.
\newblock In {\em 2018 {USENIX} Annual Technical Conference ({USENIX} {ATC}
  18)}, pages 731--743, Boston, MA, July 2018. {USENIX} Association.

\end{thebibliography}

\appendix
\section{Appendix}
\subsection{State Space}
\label{app:state_space}

We design our state space to contain aggregate network statistics obtained from ACKs within each 100ms window (the environment step-time) and an action history of length \(h\). An optimal value of \(h\) is chosen by hyperparameter sweeps to be 20. When an ACK or a packet loss notification arrives, we gather the following 20 network statistics:

\begin{center}
  \centering
  \begin{tabular}{rll}
    \toprule
    & Feature     & Description \\
    \midrule
    1 & lrtt & Latest round-trip time (RTT) in milliseconds (ms) \\
    2 & rtt\_min & Minimum RTT (in ms) since beginning of episode \\
    3 & srtt & Smoothed RTT (in ms) \\
    4 & rtt\_standing & Min RTT (in ms) over window of size \(\textit{srtt}/2\) \cite{copa} \\
    5 & rtt\_var & Variance in RTT (in ms) as observed by the network protocol \\
    6 & delay & Queuing delay measured as \(\textit{rtt\_standing} - \textit{rtt\_min}\) \\
    7 & cwnd\_bytes & Congestion window in bytes calculated as \(\cwnd * \textit{MSS}\) \\
    8 & inflight\_bytes & Number of bytes sent but unacknowledged (cannot exceed \textit{cwnd\_bytes}) \\
    9 & writable\_bytes & Number of writable bytes measured as \(\textit{cwnd\_bytes} - \textit{inflight\_bytes}\) \\
    10 & sent\_bytes & Number of bytes sent since last ACK \\
    11 & received\_bytes & Number of bytes received since last ACK \\
    12 & rtx\_bytes & Number of bytes re-transmitted since last ACK \\
    13 & acked\_bytes & Number of bytes acknowledged in this ACK \\
    14 & lost\_bytes & Number of bytes lost in this loss notification \\
    15 & throughput & Instantaneous throughput estimated from recent ACKs \\
    16 & rtx\_count & Number of packets re-transmitted since last ACK \\
    17 & timeout\_based\_rtx\_count & Number of re-transmissions due to Probe Timeout (PTO) since last ACK \\
    18 & pto\_count & Number of times packet loss timer fired before receiving an ACK \\
    19 & total\_pto\_count & Total number of times packet loss timer fired since last ACK \\
    20 & persistent\_congestion & Flag indicating whether persistent congestion is detected by the protocol \\
    \bottomrule
  \end{tabular}
\end{center}

Note that we currently compute $rtt\_min$ from the beginning of an episode, because computing it over a fixed window can cause issues during training.
Indeed, if $rtt\_min$ was the minimum RTT value observed only in the past $N$ seconds,
and the agent's actions caused significant congestion for a duration longer than $N$ seconds, then $rtt\_min$ would increase due to the congestion.
Since the congestion delay is estimated with $delay = rtt\_standing - rtt\_min$, this delay would end up being largely under-estimated, potentially leading to high rewards in spite of the ongoing congestion.
Keep in mind, however, that at deployment time $rtt\_min$ should be updated on a more frequent basis, since network conditions may be non stationary:
keeping the same value of $rtt\_min$ over a long period of time may lead the agent to over-estimate the congestion delay when higher RTTs are observed, since this might be caused by other factors independent of network congestion.

The throughput computation is an important component since it is used in the reward computation in addition to the state.
It is based on maintaining a rolling window of the last $M=10$ ACK packets: we then estimate the throughput as the total number of bytes acknowledged in this window, divided by the window duration.
To keep this throughput estimation stable in the presence of ACK bursts, the minimum window duration is 100ms.
When no ACK is received for a duration longer than the current window duration $T$, we also start linearly decreasing the estimated throughput (so that it reaches zero after $2T$ without any ACK).

As a way of normalization, the time-based features (features 1--6) are scaled by $10^{-3}$ and byte-based ones (features 7--14) are scaled by $10^{-4}$. Since there could be varying number of ACKs within any 100ms time window corresponding to an environment step, we obtain a fixed-size state vector by computing aggregate statistics from all ACKs within a window. This is achieved by computing sum, mean, standard deviation, min and max over each of the 20 features and flattening the resulting 100 features into a single vector. For features 1--9, it doesn't make sense to compute sums and we set those entries to zero in the state vector.

The history at time \(u\), \(u < t\), is encoded as a one-hot vector of the action \(a_u\) along with the congestion window as a result of that action \(\cwnd_u\). To obtain the aggregate features, we concatenate a history of length \(h\), resulting in our final state vector of length \(100 + h \times (|A| + 1)\), where \(|A|\) is the size of the action space. In our experiments, \(h = 16\) and \(|A| = 5\), resulting in a state vector of length 196.

We have also experimented with providing the index of the current training scenario as input (to the critic only).
So far this has not led to meaningful improvements but, more generally, feeding additional privileged information to the critic regarding the current network conditions may prove useful in the future, as we add more randomized scenarios.

\subsection{Action Space}
\label{app:action_space}

While there are various possible choices for the action space \(A\), we chose the simplest set of discrete updates common in conventional congestion control methods:
\[A = \{\cwnd, \cwnd/2, \cwnd-10, \cwnd+10, \cwnd\times 2\}\]
where \cwnd is the congestion window in units of Maximum Segment Size ($\textit{MSS}$). The network environment starts with an initial \cwnd of 10, and bounded updates are applied according to the policy as follows:
\[\cwnd_{t+1} = \mathrm{clip}(\mathrm{update}(\cwnd_t, a_t, A), 2, 2000)\]

where \(a_t\) is the action according to the policy at time \(t\) and is an index into the action space \(A\), \(\mathrm{update}(\cwnd_t, a_t, A)\) is a function that updates the current congestion window according to \(a_t\), and the function \(\mathrm{clip}(x, \textit{low}, \textit{high}) = \min(\max(x, \textit{low}), \textit{high})\) bounds the congestion window to reasonable limits.

\mvfstrl supports configuring the action space to any discrete set of updates easily via a simple format. The above mentioned action space can be configured as \texttt{"0,/2,-10,+10,*2"}.

\end{document}